\documentclass[letterpaper]{article} 
\usepackage{aaai2026}  
\usepackage{times}  
\usepackage{helvet}  
\usepackage{courier}  
\usepackage[hyphens]{url}  
\usepackage{graphicx} 
\usepackage{natbib}  
\usepackage{caption} 

\usepackage{algorithm}
\usepackage{algorithmic}
\usepackage{amsmath,amssymb,amsfonts}
\usepackage{subcaption}
\usepackage{booktabs}

\usepackage{newfloat}
\usepackage{listings}
\DeclareCaptionStyle{ruled}{labelfont=normalfont,labelsep=colon,strut=off} 
\floatstyle{ruled}
\newfloat{listing}{tb}{lst}{}
\floatname{listing}{Listing}
\title{Singular Value Decomposition (SVD) on Kronecker Adaptation \\ for Large Language Models}

\author{
    Yee Hin Chong \equalcontrib,
    Peng Qu\equalcontrib,
}
\affiliations{
    CRAFT Lab, Department of Computer Science and Technology,\\
    Tsinghua University, China.\\
    yuxuanzh23@mails.tsinghua.edu.cn, qp2018@mail.tsinghua.edu.cn
}

\usepackage{bibentry}

\begin{document}

\maketitle

\begin{abstract}
Large pre-trained Transformer models achieve state-of-the-art results across diverse language and reasoning tasks, but full fine-tuning incurs substantial storage, memory, and computational overhead. Parameter-efficient fine-tuning (PEFT) methods mitigate these costs by learning only a small subset of task-specific parameters, yet existing approaches either introduce inference-time latency (adapter modules), suffer from suboptimal convergence (randomly initialized low-rank updates), or rely on fixed rank choices that may not match task complexity (Kronecker-based decompositions). 

We propose \textbf{SoKA} (\textbf{S}VD \textbf{o}n \textbf{K}ronecker \textbf{A}daptation), a novel PEFT strategy that combines Kronecker-product tensor factorization with SVD-driven initialization and spectrum-aware dynamic rank selection. Our Kronecker‐Product SVD (KPSVD) procedure extracts principal components of the full weight update into compact Kronecker factors, while an adaptive rank selection algorithm uses energy-threshold and elbow-point criteria to prune negligible components.  

Empirical evaluation on LLaMA2‑7B across arithmetic reasoning (GSM8K), formal mathematics (MATH), and code generation (MBPP) demonstrates that SoKA requires only 0.99 M trainable parameters, 25\% fewer than LoRA/PiSSA, while matching or exceeding baseline performance. Moreover, SoKA exhibits faster convergence and more stable gradients, highlighting its robustness and efficiency for large-scale model adaptation.

\end{abstract}

\section{Introduction}

The rapid growth of large pre-trained Transformer models has led to unprecedented gains across a variety of language understanding and generation tasks. However, full fine-tuning of these models incurs prohibitive storage, memory, and computational costs, as each downstream task requires a separate copy of the model parameters. Parameter–efficient fine–tuning (PEFT) has emerged as a promising alternative: by freezing the bulk of the pre-trained weights and learning only a small set of task-specific parameters, PEFT methods drastically reduce the resource footprint while retaining competitive performance.

Existing PEFT approaches span a spectrum of structural priors. Adapter modules introduce small bottleneck layers into each Transformer block \cite{houlsby19a}, but add inference-time latency; Low-Rank Adaptation (LoRA) reparameterizes weight updates via low-rank factors that can be merged at inference time \cite{hu2022lora}, yet may suffer from suboptimal convergence when initialized randomly; PiSSA improves upon LoRA via Singular-Value-Decomposition (SVD)-based initialization, accelerating convergence at the cost of retaining the same rank bottleneck \cite{meng2024pissa}. More recent Kronecker-based methods \cite{braga-etal-2024-adakron, edalati2022krona, mahabadi2021compacter, yeh2024navigating} leverage structured tensor decompositions to further compress the update space and achieve high throughput, but often rely on fixed rank choices and lack spectrum-aware adaptation.

In this paper, we introduce \textbf{SoKA} (\textbf{S}VD \textbf{o}n \textbf{K}ronecker \textbf{A}daptation), a novel PEFT strategy that unifies two complementary strategies: (1) an SVD-driven initialization of Kronecker factors, which captures the principal components of the full-matrix update; and (2) a spectrum‐driven dynamic rank selection mechanism, which prunes away negligible components based on cumulative energy and elbow‐point criteria. Through the Kronecker‐Product SVD (KPSVD) formalism (Alg.~\ref{algo:kpsvd}) \cite{KPSVD}, SoKA represents each weight update as a weighted sum of Kronecker products,
\[
\Delta W \;\approx\; \sum_{k=1}^r \sigma_k\,U_k \otimes V_k,
\]
enabling efficient matrix–vector multiplication and a compact parameter footprint of only $r\,(mn + pq + 1)$ per block. The dynamic rank selection automatically adapts $r$ to the intrinsic complexity of each task, ensuring a balance between expressiveness and efficiency.

We evaluate SoKA on the LLaMA2‑7B backbone over arithmetic reasoning (GSM8K), formal mathematics (MATH), and code generation (MBPP). As shown in Table~\ref{tab:llama2}, SoKA uses only 0.99 M trainable parameters, 25\% fewer than PiSSA, while matching or exceeding their performance. Further investigations demonstrate that SoKA converges faster and with more stable gradients than PiSSA, indicating a more robust optimization landscape.

Our contributions are threefold:
\begin{itemize}
  \item We propose \textbf{KPSVD}, a Kronecker-Product SVD procedure to initialize structured adapters from the principal components of the full weight update.
  \item We develop a \textbf{dynamic rank selection} algorithm that jointly leverages energy-threshold and elbow-point criteria to adaptively determine the effective rank per layer.
  \item We show that SoKA achieves \textbf{state-of-the-art parameter efficiency} on LLama2-7B, reducing parameter count by up to 180× while maintaining or surpassing the performance of competitive PEFT baselines.
\end{itemize}
\section{Related Works}

To overcome the prohibitive costs of full model fine-tuning, parameter-efficient fine-tuning (PEFT) has been introduced as a practical and scalable paradigm for adapting large pre-trained models to a wide range of downstream tasks. Rather than updating all parameters, PEFT methods freeze the original model weights and introduce a compact set of task-specific trainable components. These additional modules are typically several orders of magnitude smaller than the full model, enabling efficient adaptation with minimal memory footprint and computational overhead. This design not only reduces the resource burden during training and inference but also facilitates multi-task deployment by allowing different tasks to share the same backbone while maintaining separate, lightweight adapters for each task.

Early implementations of PEFT centered around adapter modules, which introduce lightweight "bottleneck" modules into each Transformer layer, typically consisting of a down-projection to a low-dimensional subspace followed by a non-linearity and an up-projection back to the original dimension \cite{houlsby19a, he2022towards}. By updating only these modules during training, adapter-based methods achieve competitive performance relative to full fine-tuning while drastically reducing the number of trainable parameters. Nevertheless, the presence of additional layers in the forward pass incurs non-trivial inference overhead, which may be undesirable in latency-sensitive applications.

To address this issue, Low-Rank Adaptation (LoRA) was proposed as a more computationally efficient alternative \cite{hu2022lora}. LoRA expresses the weight update $\Delta W$ as a product of two low-rank matrices $A \in \mathbb{R}^{m \times r}$ and $B \in \mathbb{R}^{r \times n}$, while the original weight $W$ remains fixed, as illustrated in Fig \ref{fig:lora}. Only $A$ and $B$ are updated during fine-tuning, and their product can be merged into $W$ at inference time, thereby restoring the original architecture without introducing any additional inference-time latency. This reparameterization capability, together with the simplicity of low-rank matrix multiplication, has made LoRA a widely used PEFT baseline.

Despite its success, LoRA’s default practice of initializing $A$ and $B$ with random weights can result in suboptimal convergence behavior. In particular, the learning dynamics may suffer due to poor alignment between the initialized subspace and the intrinsic structure of the weight update. To address this limitation, the Principal Singular-value and Singular-vector Adaptation (PiSSA) method leverages the top-$r$ singular components of the weight matrix $W$ to initialize $A$ and $B$ \cite{meng2024pissa}. As shown in Fig \ref{fig:pissa}, the remaining residual subspace is kept frozen. This SVD-based initialization significantly accelerates convergence and improves final performance on challenging benchmarks such as GSM8K and MMLU.

However, a key limitation of LoRA and its variants lies in the representational capacity constrained by the rank $r$ of the low-rank decomposition. Theoretical analysis suggests that capturing complex weight updates in high-capacity Transformer models may require a rank that grows proportionally with the model depth and hidden dimensions \cite{hu2025computational}. Consequently, there exists an inherent trade-off between approximation fidelity and parameter efficiency, where low-rank approximations may fail to capture the task-specific dynamics of the target weight updates fully.

To overcome this bottleneck, researchers have proposed a family of Kronecker product-based PEFT methods that offer a richer structural prior for modeling weight updates. For instance, \cite{mahabadi2021compacter} parameterizes weight updates as a sum of Kronecker products between shared “slow” global factors and per-layer “fast” rank-one factors. This formulation achieves state-of-the-art results on GLUE and SuperGLUE while tuning fewer than 0.1\% of the total parameters. Notably, it introduces minimal runtime overhead due to its structured parameterization. Building on this idea, \cite{edalati2022krona} and its adaptive variant \cite{braga-etal-2024-adakron}, completely replace low-rank projection matrices with Kronecker-structured components. These models further improve parameter efficiency and inference-time speed while maintaining strong accuracy across a range of NLP benchmarks. More recently, \cite{yeh2024navigating} reinterprets LoRA within the Kronecker product framework, achieving a highly compressed and computationally efficient PEFT method that retains the flexibility of LoRA.
\section{SoKA: SVD on Kronecker Adaptation}

\begin{figure*}[htbp]
    \centering
    \begin{subfigure}[t]{0.32\textwidth}
        \centering
        \includegraphics[trim=300 130 300 130, clip, width=\linewidth]{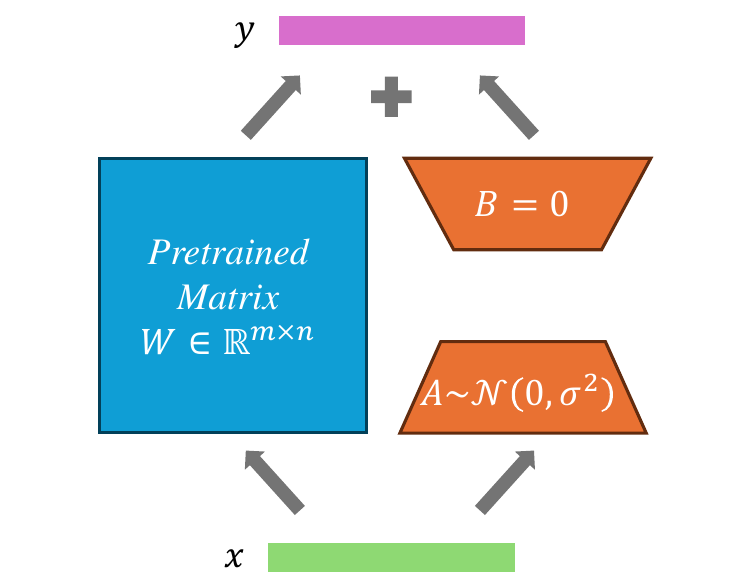}
        \caption{LoRA}
        \label{fig:lora}
    \end{subfigure}
    \hfill
    \begin{subfigure}[t]{0.32\textwidth}
        \centering
        \includegraphics[trim=300 130 300 130, clip, width=\linewidth]{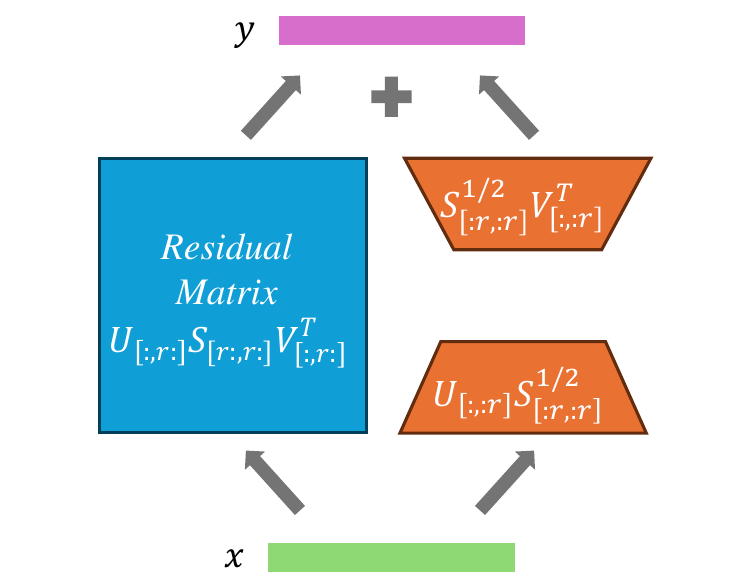}
        \caption{PiSSA}
        \label{fig:pissa}
    \end{subfigure}
    \hfill
    \begin{subfigure}[t]{0.32\textwidth}
        \centering
        \includegraphics[trim=300 130 300 130, clip, width=\linewidth]{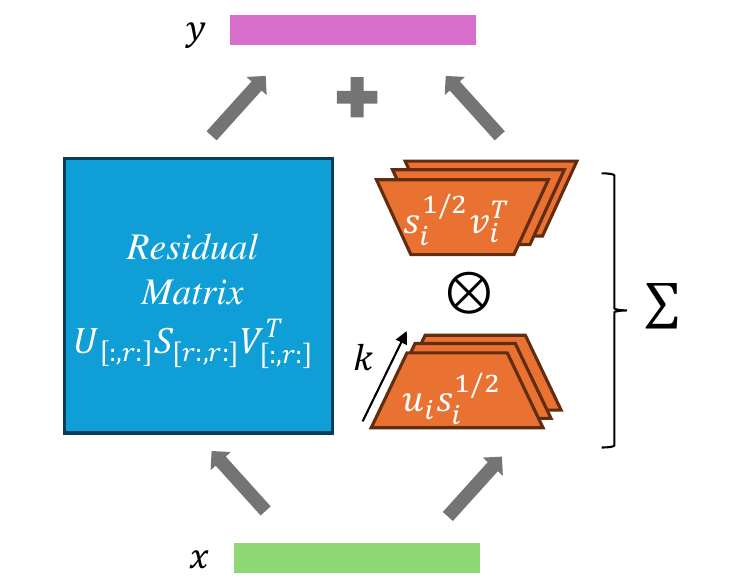}
        \caption{SoKA (Ours)}
        \label{fig:soka}
    \end{subfigure}
    \caption{Visualization of LoRA, PiSSA, and SoKA.}
    \label{fig:adapters}
\end{figure*}

In this work, we propose a novel integration of two complementary strategies: the use of SVD-based principal component initialization and Kronecker product decomposition \cite{KPSVD}. Our method, SoKA, as illustrated in Fig \ref{fig:soka}, formulates the weight update $\Delta W$ as a sum of structured Kronecker products.

\subsection{Kronecker‐Product SVD for Adapter Initialization}

The Kronecker product of two matrices \(A\in\mathbb{R}^{m\times n}\) and \(B\in\mathbb{R}^{p\times q}\), denoted \(A\otimes B\), is the block matrix.
\[
A\otimes B \;=\;
\begin{bmatrix}
a_{11}B & a_{12}B & \cdots & a_{1n}B\\
a_{21}B & a_{22}B & \cdots & a_{2n}B\\
\vdots  & \vdots  & \ddots & \vdots \\
a_{m1}B & a_{m2}B & \cdots & a_{mn}B
\end{bmatrix}
\;\in\;\mathbb{R}^{(m p)\times(n q)}.
\]

Given a large square weight matrix \(W\in\mathbb{R}^{mn\times pq}\) with \(N=mn=pq\), we first reshape
\[
W' = \operatorname{reshape}(W,\,(mn)\times(pq))\in\mathbb{R}^{N\times N}
\]
and compute its (truncated) singular value decomposition
\[
W' \;=\;\sum_{k=1}^{N} \sigma_k\,x_k\,y_k^\top
\quad\Longrightarrow\quad
W_r \approx \sum_{k=1}^r \sigma_k\,x_k\,y_k^\top.
\]
Then each singular vector \(x_k\in\mathbb{R}^{mn}\) and \(y_k\in\mathbb{R}^{pq}\) is 'unvectorized' into smaller factors.
\[
U_k = \operatorname{reshape}(x_k,\,(m,n)), 
\quad
V_k = \operatorname{reshape}(y_k,\,(p,q)),
\]
yielding the Kronecker‐product approximation
\[
W \;\approx\; \sum_{k=1}^r \sigma_k\,U_k \otimes V_k.
\]

This process can be described by Algorithm~\ref{algo:kpsvd}.

\begin{algorithm}
\caption{Kronecker‐Product SVD (KPSVD)}\label{algo:kpsvd}
\begin{algorithmic}[1]
  \REQUIRE \(W\in\mathbb{R}^{mn\times pq}\), integers \(m,n,p,q,r\)
  \ENSURE \(\{(\sigma_i,U_i,V_i)\}_{i=1}^r\) with \(W\approx\sum_i\sigma_i\,U_i\otimes V_i\)
  \STATE \(W'\gets \mathrm{reshape}(W,\,(mn)\times(pq))\)
  \STATE \([U,\Sigma,V^\top]\gets\mathrm{svd}(W')\)
  \STATE \(U_r\gets U[:,1\!:\!r],\;\Sigma_r\gets\Sigma[1\!:\!r],\;V_r^\top\gets V^\top[1\!:\!r,:]\)
  \FOR{\(i=1\) \TO \(r\)}
    \STATE \(\sigma_i\gets\Sigma_k[i]\)
    \STATE \(u_i\gets U_r[:,i],\;v_i\gets V_r^\top[i,:]\)
    \STATE \(U_i\gets \mathrm{reshape}(u_i,\,(m,n))\)
    \STATE \(V_i\gets \mathrm{reshape}(v_i,\,(p,q))\)
  \ENDFOR
  \RETURN \(\{(\sigma_i,U_i,V_i)\}_{i=1}^r\)
\end{algorithmic}
\end{algorithm}

For an $N\times N$ weight matrix $W$ (with $N = mp = nq$), a LoRA‐style update $\Delta W_{\mathrm{LoRA}}
\;=\; A\,B^{\mathsf T} \quad A,B\in\mathbb{R}^{N\times r_{\mathrm{lora}}}$
of rank $r_{\mathrm{lora}}$ requires storing $A$ and $B$ and incurs a dense multiplication cost of $\mathcal{O}\bigl(N\times r_{\mathrm{lora}}\times N\bigr)
=\mathcal{O}\bigl(r_{\mathrm{lora}}N^2\bigr).$

In contrast, a KPSVD‐based adapter is constructed as
\[
\Delta W_{\mathrm{KPSVD}}
\;=\;
\sum_{k=1}^{r_{\mathrm{kp}}}
\sigma_k\,\bigl(U_k \otimes V_k\bigr),
\]
which, under naive reconstruction, also involves $\mathcal{O}\bigl(r_{\mathrm{kp}}\,N^2\bigr)$ operations. 

However, each Kronecker term admits an efficient matrix–vector multiplication via
\[
\bigl(U_k\otimes V_k\bigr)\,\mathrm{vec}(X)
\;=\;
\mathrm{vec}\bigl(V_k\,X\,U_k^{\mathsf T}\bigr),
\]
reducing the per‐component cost from $\mathcal{O}(N^2)$ to $\mathcal{O}(mpq + npq)$ and yielding a total complexity of $\mathcal{O}\bigl(r_{\mathrm{kp}}\,(mpq + npq)\bigr)$.

In terms of storage, LoRA requires $2N\,r_{\mathrm{lora}}$ parameters for $A$ and $B$, whereas KPSVD only stores $r_{\mathrm{kp}}$ singular values and the factor matrices $U_k\in\mathbb{R}^{m\times n}$ and $V_k\in\mathbb{R}^{p\times q}$, totaling $r_{\mathrm{kp}}\,(mn + pq + 1)$ parameters, often substantially fewer than LoRA’s parameter footprint for typical block sizes.

Moreover, each Kronecker term can be computed independently and then element‑wise reduced (e.g., summed), fully leveraging hardware parallelism and yielding high‑performance kernels. As a result, KPSVD‑based adapters achieve substantial speedups in large‑scale fine‑tuning while preserving structured low‑rank efficiency.

\subsection{Dynamic Rank Selection}

Dynamic rank selection enables each adapter to adjust its representational capacity to the actual complexity of the downstream task. After performing KPSVD on the adapter’s weight update, we obtain a nonincreasing sequence of singular values  
\[
S = [\sigma_1 \ge \sigma_2 \ge \dots \ge \sigma_{\min(m,n)}].
\]  
The rate of decay in this spectrum reveals the intrinsic dimensionality of the update: a sharp drop indicates that only a few leading directions carry most of the signal, whereas a long tail implies richer structure. By examining both the cumulative energy and the “elbow” in the spectrum, we can determine the smallest rank \(r\) that retains nearly all informative components while discarding negligible modes that tend to introduce noise or overfitting. This adaptive mechanism ensures that each adapter dedicates precisely the capacity needed to capture task‑specific features, optimizing memory and computation without sacrificing accuracy.

Our method, SoKA, combines two greedy heuristics—an energy‑threshold test and an elbow‑point detection—to estimate the optimal rank:

\begin{enumerate}
  \item \textbf{Energy‑Threshold Criterion.} \cite{Jolliffe1986}
    Define the cumulative energy fraction  
    \[
      E(k) = \frac{\sum_{i=1}^k \sigma_i^2}{\sum_{j=1}^{\min(m,n)} \sigma_j^2}.
    \]  
    Select the smallest \(k\) such that \(E(k)\ge\tau\), where \(\tau\in(0,1)\) is a user‑specified threshold (e.g.\ \(\tau=0.90\) or \(0.95\)). 
  \item \textbf{Elbow‑Point Criterion.} \cite{Jackson1993}
    Compute the successive gaps  
    \[
      \delta_i = \sigma_i - \sigma_{i+1}, 
      \quad
      i^* = \arg\max_i \delta_i,
    \]  
    and set \(r_{\mathrm{elbow}} = i^*\).
\end{enumerate}

Let  
\[
r_{\mathrm{energy}} = \min\{\,k : E(k)\ge\tau\}, 
\qquad 
r_{\mathrm{elbow}} = i^*.
\]  
We then define the working rank as  
\[
r = \min\bigl(r_{\mathrm{energy}},\,r_{\mathrm{elbow}}\bigr),
\]  
optionally constrained within user bounds \(r_{\min}\le r\le r_{\max}\).

This enables each adapter to autonomously adjust its capacity by first exploiting KPSVD and then applying spectrum‐driven pruning. This two-stage reduction, structural factorization followed by principal rank selection, yields a lean yet expressive adaptation mechanism. By compressing the tunable parameter space while retaining essential task‐specific features, our method offers a compact, flexible, and scalable solution for finetuning under limited computational budgets.

\section{Experiments}

We evaluate SoKA against standard LoRA, PiSSA, and full fine-tuning baselines on the LLaMA2‑7B model across a diverse set of tasks, including arithmetic reasoning (GSM8K) \cite{cobbe2021gsm8k}, formal mathematics (MATH) \cite{hendrycks2021math}, and code generation (MBPP) \cite{austin2021mbpp}. As summarized in Table~\ref{tab:llama2}, SoKA introduces only 0.99 million trainable parameters, representing a reduction of approximately 25\% compared to both LoRA and PiSSA configurations (1.33M), and a dramatic decrease of over 180× compared to full fine-tuning (184M). This substantial parameter reduction highlights SoKA's efficiency in adapting large language models, significantly lowering the memory and storage overhead associated with task-specific fine-tuning.

Despite its compact parameter footprint, SoKA achieves competitive performance across all benchmarks. As shown in Table~\ref{tab:llama2}, SoKA consistently and significantly outperforms standard LoRA across all tasks, demonstrating improved generalization despite using fewer parameters. Compared to PiSSA—a recent method that leverages SVD-initialized low-rank adaptation—SoKA achieves comparable results, trailing by less than 1 point on each benchmark while maintaining a more lightweight design. These results indicate that SoKA not only outperforms traditional low-rank adaptation in both efficiency and effectiveness, but also matches the performance level of more complex methods like PiSSA, despite using fewer trainable parameters and a simpler structure.

\begin{figure}[t]
    \centering
    \begin{subfigure}[t]{0.9\columnwidth}
        \centering
        \includegraphics[trim=15 0 45 30, clip, width=\linewidth]{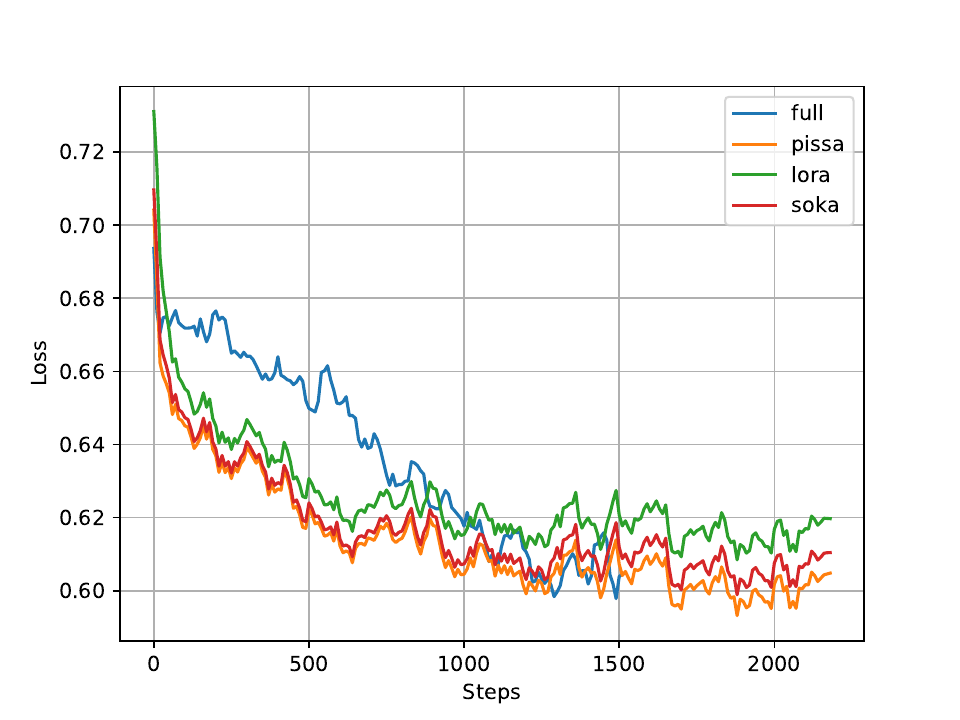}
        \caption{Loss over steps}
        \label{fig:loss}
    \end{subfigure}
    
    \vspace{0.5em}

    \begin{subfigure}[t]{0.9\columnwidth}
        \centering
        \includegraphics[trim=20 0 45 30, clip, width=\linewidth]{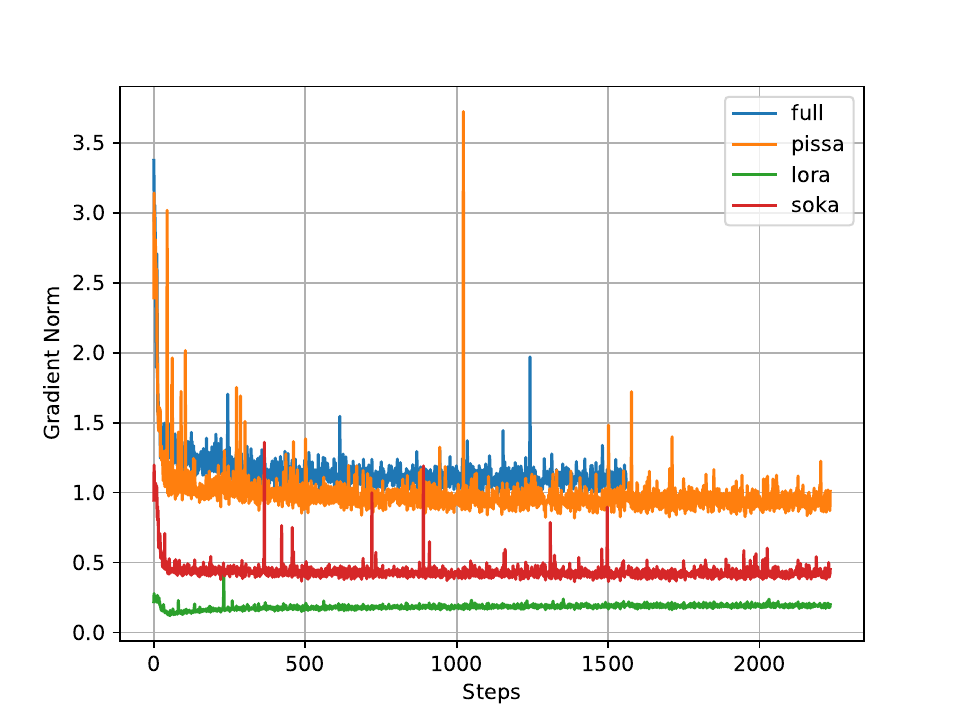}
        \caption{Gradient norm over steps}
        \label{fig:grad_norm}
    \end{subfigure}
    
    \caption{Training dynamics: (a) loss and (b) gradient norm over the training steps of LoRA (indicated in green), PiSSA (in orange), SoKA (in red), and full parameter fine-tuning (in blue).}
    \label{fig:training_dynamics}
\end{figure}

Beyond final performance metrics, we also analyze the training dynamics of SoKA to better understand its optimization behavior. Figures~\ref{fig:loss} and~\ref{fig:grad_norm} visualize the loss curves and gradient norms throughout training. SoKA exhibits faster and smoother convergence, achieving lower loss values than PiSSA after the initial 500 steps and maintaining this advantage throughout the training process. In contrast, PiSSA’s training curve reveals higher variance and slower convergence, indicating less stable optimization.

In terms of gradient behavior, SoKA demonstrates remarkably stable and restrained gradient norms, rarely exceeding a magnitude of 1.5. This contrasts with PiSSA, which exhibits pronounced fluctuations and peaks around 3.0. The smooth gradient flow of SoKA suggests greater training robustness, as it mitigates risks such as gradient explosion and erratic updates. The combination of stable gradients and efficient convergence indicates that SoKA is not only parameter-efficient, but also optimizer-friendly, enabling robust and efficient fine-tuning on large-scale models.

\begin{table*}[htbp]
  \caption{Comparison of PEFT methods LLaMA2-7B.}
  \label{tab:llama2}
  \centering
  \begin{tabular}{lcccc}
    \toprule
    \textbf{Strategy} & \textbf{\#Params} & \textbf{GSM8K} & \textbf{MATH} & \textbf{MBPP} \\
    \midrule
    Full FT & 184M & $60.42 \pm 0.21$ & $12.6 \pm 0.22$ & $42.6 \pm 0.25$ \\
    LoRA ($r_\text{lora} = 128$) & 1.33M & $44.58 \pm 0.12$ & $6.12 \pm 0.33$ & $35.5 \pm 0.14$ \\
    PiSSA ($r_\text{lora} = 128$) & 1.33M & $53.37 \pm 0.55$ & $8.17 \pm 0.34$ & $40.4 \pm 0.63$ \\
    SoKA ($r_\text{soka} \leq 128$) & 0.99M & $52.19 \pm 0.32$ & $7.93 \pm 0.33$ & $39.5 \pm 0.32$ \\
    \bottomrule
  \end{tabular}
\end{table*}

\section{Conclusion and Future Work}

We introduced \textbf{SoKA}, a PEFT strategy combining Kronecker‐Product SVD and dynamic rank selection to capture principal weight updates in compact form. On LLaMA2‑7B (GSM8K, MATH, MBPP), SoKA uses just 0.99 M parameters, 25\% fewer than LoRA/PiSSA, while matching or surpassing their performance, and converges faster with more stable gradients.

Future directions include:
\begin{itemize}
  \item Incorporating \textbf{structured sparsity} or \textbf{low‑bit quantization} into Kronecker factors to further cut resource use.
  \item Exploring \textbf{alternative bottlenecks} (e.g., tensor‑train or block‐diagonal) and \textbf{automated spectrum tuning} for even finer adaptation.
\end{itemize}

\bibliography{main}

\begin{thebibliography}{15}
\providecommand{\natexlab}[1]{#1}

\bibitem[{Austin et~al.(2021)Austin, Odena, Nye, Bosma, Michalewski, Dohan, Jiang, Cai, Terry, Le, and Sutton}]{austin2021mbpp}
Austin, J.; Odena, A.; Nye, M.; Bosma, M.; Michalewski, H.; Dohan, D.; Jiang, E.; Cai, C.; Terry, M.; Le, Q.; and Sutton, C. 2021.
\newblock Program Synthesis with Large Language Models.
\newblock arXiv:2108.07732.

\bibitem[{Batselier and Wong(2017)}]{KPSVD}
Batselier, K.; and Wong, N. 2017.
\newblock A constructive arbitrary-degree Kronecker product decomposition of tensors.
\newblock \emph{Numerical Linear Algebra with Applications}, 24(5): e2097.
\newblock E2097 nla.2097.

\bibitem[{Braga, Raganato, and Pasi(2024)}]{braga-etal-2024-adakron}
Braga, M.; Raganato, A.; and Pasi, G. 2024.
\newblock {A}da{K}ron: An Adapter-based Parameter Efficient Model Tuning with Kronecker Product.
\newblock In Calzolari, N.; Kan, M.-Y.; Hoste, V.; Lenci, A.; Sakti, S.; and Xue, N., eds., \emph{Proceedings of the 2024 Joint International Conference on Computational Linguistics, Language Resources and Evaluation (LREC-COLING 2024)}, 350--357. Torino, Italia: ELRA and ICCL.

\bibitem[{Cobbe et~al.(2021)Cobbe, Kosaraju, Bavarian, Chen, Jun, Kaiser, Plappert, Tworek, Hilton, Nakano, Hesse, and Schulman}]{cobbe2021gsm8k}
Cobbe, K.; Kosaraju, V.; Bavarian, M.; Chen, M.; Jun, H.; Kaiser, L.; Plappert, M.; Tworek, J.; Hilton, J.; Nakano, R.; Hesse, C.; and Schulman, J. 2021.
\newblock Training Verifiers to Solve Math Word Problems.
\newblock arXiv:2110.14168.

\bibitem[{Edalati et~al.(2022)Edalati, Tahaei, Kobyzev, Nia, Clark, and Rezagholizadeh}]{edalati2022krona}
Edalati, A.; Tahaei, M.; Kobyzev, I.; Nia, V.~P.; Clark, J.~J.; and Rezagholizadeh, M. 2022.
\newblock KronA: Parameter Efficient Tuning with Kronecker Adapter.
\newblock arXiv:2212.10650.

\bibitem[{He et~al.(2022)He, Zhou, Ma, Berg-Kirkpatrick, and Neubig}]{he2022towards}
He, J.; Zhou, C.; Ma, X.; Berg-Kirkpatrick, T.; and Neubig, G. 2022.
\newblock Towards a Unified View of Parameter-Efficient Transfer Learning.
\newblock In \emph{International Conference on Learning Representations}.

\bibitem[{Hendrycks et~al.(2021)Hendrycks, Burns, Kadavath, Arora, Basart, Tang, Song, and Steinhardt}]{hendrycks2021math}
Hendrycks, D.; Burns, C.; Kadavath, S.; Arora, A.; Basart, S.; Tang, E.; Song, D.; and Steinhardt, J. 2021.
\newblock Measuring Mathematical Problem Solving With the MATH Dataset.
\newblock arXiv:2103.03874.

\bibitem[{Houlsby et~al.(2019)Houlsby, Giurgiu, Jastrzebski, Morrone, De~Laroussilhe, Gesmundo, Attariyan, and Gelly}]{houlsby19a}
Houlsby, N.; Giurgiu, A.; Jastrzebski, S.; Morrone, B.; De~Laroussilhe, Q.; Gesmundo, A.; Attariyan, M.; and Gelly, S. 2019.
\newblock Parameter-Efficient Transfer Learning for {NLP}.
\newblock In Chaudhuri, K.; and Salakhutdinov, R., eds., \emph{Proceedings of the 36th International Conference on Machine Learning}, volume~97 of \emph{Proceedings of Machine Learning Research}, 2790--2799. PMLR.

\bibitem[{Hu et~al.(2022)Hu, yelong shen, Wallis, Allen-Zhu, Li, Wang, Wang, and Chen}]{hu2022lora}
Hu, E.~J.; yelong shen; Wallis, P.; Allen-Zhu, Z.; Li, Y.; Wang, S.; Wang, L.; and Chen, W. 2022.
\newblock Lo{RA}: Low-Rank Adaptation of Large Language Models.
\newblock In \emph{International Conference on Learning Representations}.

\bibitem[{Hu et~al.(2025)Hu, Su, jui kuo, Song, and Liu}]{hu2025computational}
Hu, J. Y.-C.; Su, M.; jui kuo, E.; Song, Z.; and Liu, H. 2025.
\newblock Computational Limits of Low-Rank Adaptation (Lo{RA}) Fine-Tuning for Transformer Models.
\newblock In \emph{The Thirteenth International Conference on Learning Representations}.

\bibitem[{Jackson(1993)}]{Jackson1993}
Jackson, D.~A. 1993.
\newblock Stopping Rules in Principal Components Analysis: A Comparison of Heuristical and Statistical Approaches.
\newblock \emph{Ecology}, 74(8): 2204–2214.

\bibitem[{Jolliffe(1986)}]{Jolliffe1986}
Jolliffe, I.~T. 1986.
\newblock \emph{Principal Component Analysis}.
\newblock Springer New York.
\newblock ISBN 9781475719048.

\bibitem[{mahabadi, Henderson, and Ruder(2021)}]{mahabadi2021compacter}
mahabadi, R.~K.; Henderson, J.; and Ruder, S. 2021.
\newblock Compacter: Efficient Low-Rank Hypercomplex Adapter Layers.
\newblock In Beygelzimer, A.; Dauphin, Y.; Liang, P.; and Vaughan, J.~W., eds., \emph{Advances in Neural Information Processing Systems}.

\bibitem[{Meng, Wang, and Zhang(2024)}]{meng2024pissa}
Meng, F.; Wang, Z.; and Zhang, M. 2024.
\newblock Pi{SSA}: Principal Singular Values and Singular Vectors Adaptation of Large Language Models.
\newblock In \emph{The Thirty-eighth Annual Conference on Neural Information Processing Systems}.

\bibitem[{Yeh et~al.(2024)Yeh, Hsieh, Gao, Yang, Oh, and Gong}]{yeh2024navigating}
Yeh, S.-Y.; Hsieh, Y.-G.; Gao, Z.; Yang, B. B.~W.; Oh, G.; and Gong, Y. 2024.
\newblock Navigating Text-To-Image Customization: From Ly{CORIS} Fine-Tuning to Model Evaluation.
\newblock In \emph{The Twelfth International Conference on Learning Representations}.

\end{thebibliography}

\end{document}